# Comparative study and enhancement of Camera Tampering Detection algorithms

Mabrouka Hagui[1], Mohamed Ali Mahjoub[1] and Ahmed Boukhris[1]
[1]SAGE Research Unit
ENISo School of Engineers of Sousse
University of Sousse, Tunisia
{mabrouka.hagui@gmail.com - medali.mahjoub@ipeim.rnu.tn}

**Abstract**

*Recently the use of video surveillance systems is widely increasing. Different places are equipped by camera surveillances such as hospitals, schools, airports, museums and military places in order to ensure the safety and security of the persons and their property. Therefore it becomes significant to guarantee the proper working of these systems. Intelligent video surveillance systems equipped by sophisticated digital camera can analyze video information's and automatically detect doubtful actions. The camera tampering detection algorithms may indicate that accidental or suspicious activities have occurred and that causes the abnormality works of the video surveillance.*
*Camera Tampering Detection uses several techniques based on image processing and computer vision. In this paper, comparative study of performance of three algorithms that can detect abnormal disturbance for video surveillance is presented.*

*Keywords*---**Camera Tampering detection, image processing, constraints, effective results.**

## 1. Introduction

Safety and prevention against accidental or malicious risks are an extreme necessity for many places and areas requiring security measures such as video surveillance used for several years by companies and public agencies, including banks, in order to protect their property and persons therein. Today, computer vision enables artificial intelligence to analyze an image and to provide current information about the scene.
For this reason, research aims to make the video surveillance more intelligent and able to prevent threats that cause the malfunction of video surveillance. This requires the capture of camera tampering attempts.

In the literature many of the works that have been presented exploit visual information for detecting camera tampering. Some of these methods are based on entropy value, background subtraction, DCT, FFT and contour of image. The purpose of the present paper is to conduct a comparative study between several algorithms used to detect camera sabotage in order to design a new method which resolve the anomalies of the previous algorithms.

Next, we will begin with a description of common example of the camera tampering in section 2. Section 3 presents an investigation of different features used to detect suspicious events. In section 4, we present some algorithms used to detect camera tampering. Section 5 exposes the experiments results. Section 6 summarizes our results and discusses the possible topics for future works.

## 2. Camera Tampering

In this paper, the camera tampering will be considered as any sustained event that can alter the working of video surveillance system. The common examples of this doubtful activity include camera occlusion, camera defocus and camera motion.

### 2.1. Camera occlusion

The obstruction or occlusion of the camera is a malicious action of preventing the camera to capture the current scene which causes partial or total loss of vision. It may occur by placing an opaque object in front of the camera to cover its lens.

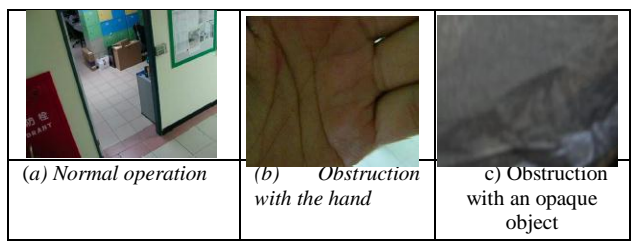

(*a*) *Normal operation* | (*b*) *Obstruction with the hand* | c) Obstruction with an opaque object

Figure 1 Camera Occulusion

## 2.2. Defocus of the camera

Defocus of the camera causes the reduction of the visibility of the captured scene. As result of this event, the image becomes blurred and analyzing and identifying its content seems difficult.

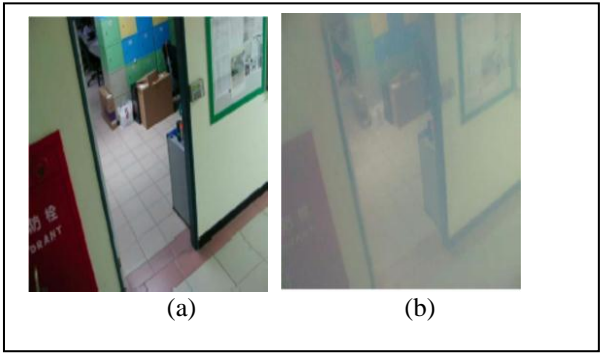

Figure 2 Camera defocus : (a) Normal vision  (b) camera defocus

This unexpected action can be caused by the damage of the device, the change of camera configuration, some weather conditions affecting cameras installed outside such as fog, water droplets... etc.

## 2.3. Camera motion

The camera tampering can be caused by the deviation of the position of the surveillance camera from its original angle. This event is called camera motion. The figure 3 shows an example of this incident.

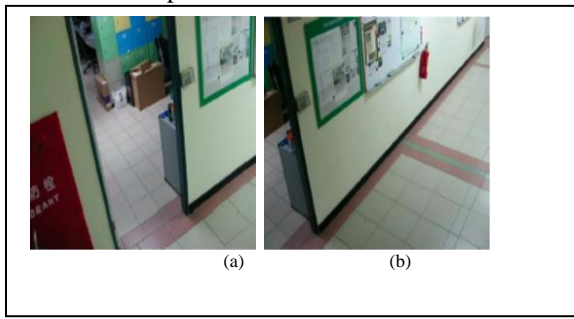

Figure 3 Camera Motion: (a) Normal vision
(b) vision after camera deviation

## 3. Camera tampering Methods

### 3.1. Camera occlusion methods

In the literature, several methods are proposed to detect the camera obstruction event.

#### 3.1.1. Method 1: calculation of entropy

Entropy is a very important quantity of information theory. It is due to Shannon and represents the average minimum amount of information to represent a digital source unambiguous. In the case of digital images, it is expressed with the following formula:

$$E = -\sum_{k} P(I_k) \log_2 [P(I_k)] \quad (1)$$

Where $P(I_K)$ represents the probability of the intensity of the pixel in level k.

This definition states that the entropy of an image is a measure of its complexity. When the image is complex and contains varieties intensity levels by the existence of multiple objects in the scene, the value of the image entropy increases and is far from zero. However, if the image is uniform and has only one color without varieties of pixel intensity levels, the entropy of the image will be substantially reduced and considerably approaches to 0.

The following figures show the difference of the entropy value between the case of the image of a normal scene and the obstruction of the camera.

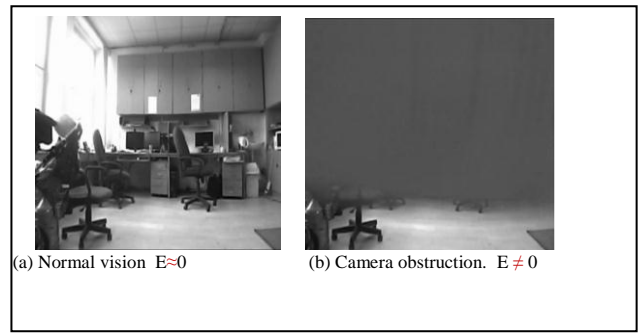

(a) Normal vision  E≈0         (b) Camera obstruction.  E ≠ 0

Figure 4 Influence on the entropy E

This approach remains true as the obstructing object is placed near the device or directly on the camera lens.

#### 3.1.2. Method 2: Histogram Analysis

In digital imaging, the histogram is a basic image processing tool that shows the distribution of intensities (gray level image or color image) of the pixels of the image.

In our case, we used the 2D image into grayscale to analyze the intensity distribution in the histogram for each image.

In normal case, the image shows the whole scene, the distribution of pixels in the histogram is distributed all over levels of intensity. However, when the camera lens is covered by an object, the histogram of the image extracted from the video at that moment is expected to have a high number of significant pixels in a specific range, because most of the scene is occupied by the levels of intensities representing the color of the hedging object that is colored or dark.

The following figure shows the distribution of the histogram for the case of a normal scene and when the camera lens is covered by the hand.

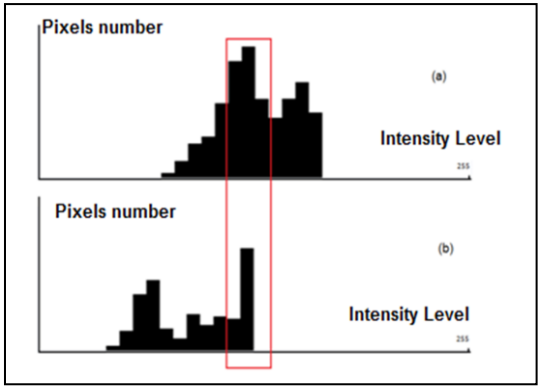

Figure 5 (a) histogram distribution in the case of obstruction (b) histogram distribution in the case of a normal scene.

### 3.1.3. Analysis of the contour and the edge of the image

This technique is based on the detection of the contour representing the objects and the edges present in the corner where the camera was positioned. Using dedicated filters for detecting the shape of the scene such as "Sobel" filter or "Canny" filter or "Perwitt" filter, we can obtain the result shown in the following figures according to the case of obstruction and the case of normal operation.

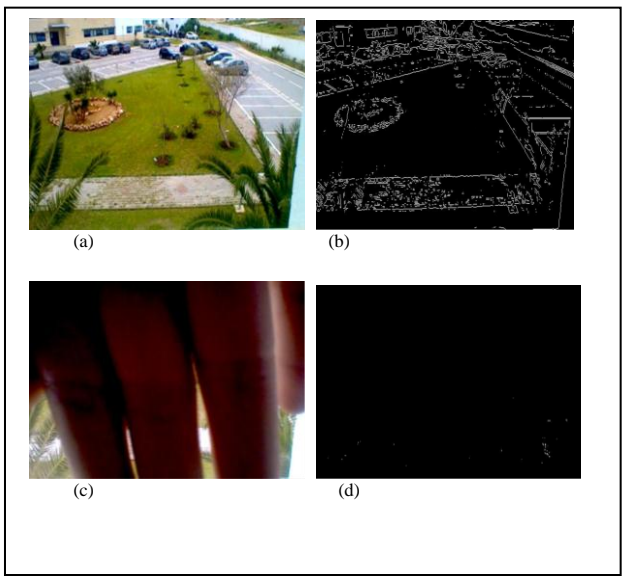

Figure 6 Influence of Obstruction of the camera on the image contour

### 3.2. Camera defocus methods

To detect if the reduction of the scene visibility, the common used methods are based on Fast Fourier Transform (FFT) and Discrete Cosine transform (DCT).

### 3.2.1. Fast Fourier Transform (FFT)

The FFT is an algorithm for calculation of the Discrete Fourier transform (DFT) and has a significant impact on the development of applications in digital signal processing. It reduces significantly the number of operations to be performed. Instead of performing $T^2$ operations for the DFT, it simply makes $T \log_2 T$ .

As an image processing tool, the FFT provides the equivalent software of a spectrum analyzer that engineers use to draw the graph of the frequencies contained in an analog signal.

This algorithm is commonly used to transform discrete time domain data in the frequency domain as indicated in equations (2) and (3):

$$f(x,y) = \sum_{u=0}^{M} \sum_{v=0}^{N} f(x,y) e^{j\,2\,\pi\,(\frac{ux}{M}+\frac{vy}{N})} \qquad (2)$$

$$F(u,v) = \frac{1}{MN} \sum_{x=0}^{M} \sum_{y=0}^{N} f(x,y) e^{-j\,2\,\pi\,(\frac{ux}{M}+\frac{vy}{N})} \quad (3)$$

In image processing, the FFT separates the frequency domain into 2 types of data:

- Coefficients of high frequencies represent the consistency and density of the edges of objects in the image.
- Coefficients of low frequencies represent the variations of pixel intensities and colours in the image.

We focus on areas of high frequencies to extract coefficients that have contours and edges of the image that are most sensitive to changes in visibility and sharpness of the captured scene.

As shown in the figure below, in case of a clear picture, contours and edges are strong and consistent, whereas in the blurred one, contours are degraded and lost.

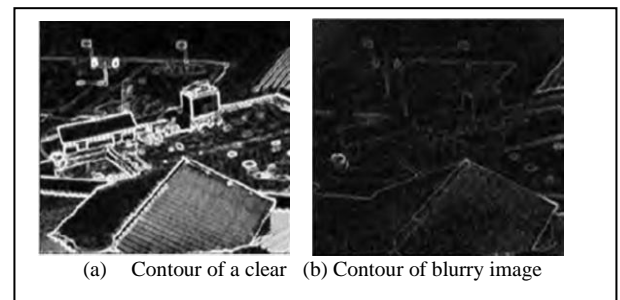

(a) Contour of a clear   (b) Contour of blurry image

Figure 7 Degradation of the contour in case of reduction of the scene's visibility

The figure 8 shows a comparison between FFT spectrum in cases of a clear image and blurred image.

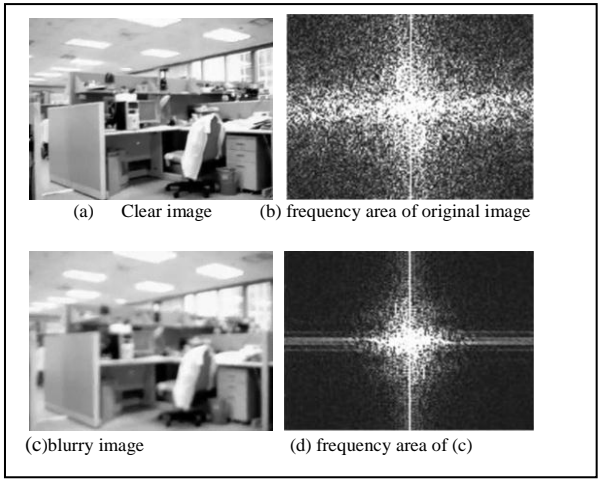

Figure 8 Comparison between FFT spectrum in cases of a clear image and blurred image.

### 3.2.2. Method 2: Discrete Cosine Transform (DCT)

This transformation is close to the FFT. The projection core is a cosine, which means that the frequency coefficients are real, unlike the FFT in which the core is a complex exponential and the coefficients are complex.

The concept is almost the same as the FFT, the application of the DCT converts the information of the image of the spatial area into the same representation in the frequency area.

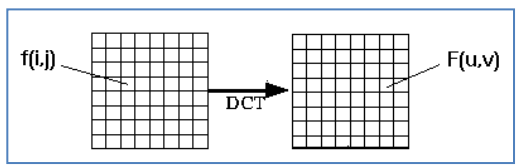

Figure 9 Image processing from time area to frequency area

The discrete cosine transform (DCT) is used to separate the image into parts (or spectral sub-bands).

Consider the case of a first level of decomposition; the DCT separates the image into 4 different main parts (relative to the visual quality of the image) as presenting in figure 10

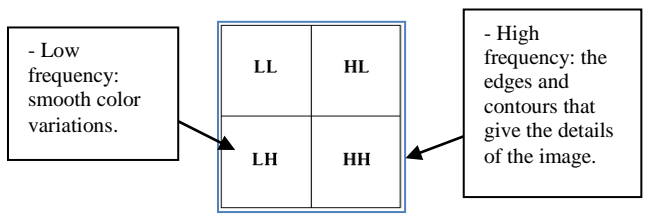

Figure 10 Decomposition first level DCT

## 3.3. Camera motion detection methods

The mainly used methods to detect a camera motion incident are the background subtraction to bring up the foreground objects of the scene and the change of the position of background pixels.

### 3.3.1. Method 1: Background subtraction

For the background subtraction, essentially two methods can be employed: the difference between successive frames
in the case of low complexity of the background or the mixture of Gaussians otherwise.

**a- Difference between successive frames**

The first method calculates the difference between pixels having the same position in successive frames. If this difference exceeds a definite threshold $\tau$, then that pixel will be considered as a pixel belonging to the foreground objects.
it uses the following equation:

$$T_D(x, y) = |T_t(x, y) - T_{t-1}(x, y)| > \tau \qquad (4)$$

(4) Where $T_D(x, y)$ corresponds to the difference between the current frame and the delayed frame in the pixel (x,y), $T_t(x, y)$ : Intensity of the pixel (x, y) of the current frame in grayscale and $T_{t-1}(x, y)$: Intensity of the pixel (x, y) of the delayed frame in grayscale

**b- Gaussian Mixture**

This method is presented in [4] can handle multiple background model distributions and provides a description of both the background and foreground. The probability of observing a certain pixel value x, at time t is described by means of a mixture of K Gaussian distributions.
This probabilistic technique allows to take into account the slight change in light intensity considered as generated noises.
The distribution (average) characterizes the color value of the pixel belonging to the background and the variance shows the variation around this value.
By using a learning rate that repairs the speed of adaptation, mean and variance parameters will be recursively updated.
The background and its variance are determined for each image, as follows:

$$F_t = \alpha\, F_{t-1} - (1 - \alpha)I_t \qquad (5)$$
$$v_t = \alpha\, v_{t-1} - (1 - \alpha)(F_t - I_t)^2 \qquad (6)$$

$F_t$ : Background image at time t
$v_t$ : Variance value at time t
$I_t$: Current image

The foreground is formed when we know the pixels belonging to the back of the scene as they are distant from the mean value. In general, the variance is used in the estimation of the foreground as follows:

$$M_t(x,y) = \begin{cases} 1, & |I_t(x,y) - F_t(x,y)| < 2{,}5\sqrt{v_t(x,y)} \\ 0, & otherwise \end{cases} \quad (7)$$

$M_t$ : Binary image

### 3.3.2. Method 2: Position of the image pixels

The concept is to perceive the change in the intensity of each pixel in its position (x, y) especially those belonging to the background. If most of the pixels intensities change this means that their position have also changed and the intelligent system detect that camera motion has occurred.

## 4. Camera Tampering detection Enhancement Methods

### 4.1. Algorithm 1: Detection of the obstruction of the camera

This method is proposed in [10] uses the variation of the Entropy value to detect the camera occlusion.
The idea is to calculate the entropy value $E_n$ of the current frame $T_n$ and to compare it with the Entropy $E_{n-1}$ of the delayed frame $T_{n-1}$. If the obstruction is so abrupt then the ratio between the two values of the entropies considerably decreases and is less than a threshold $\alpha$ or $\alpha \in [0\ 1]$.
 This condition is represented as follows:

$$\textbf{Obstruction} = \begin{cases} True, & if\ \frac{E_n}{E_{n-1}} < \alpha \\ False, & if\ \frac{E_n}{E_{n-1}} > \alpha \end{cases} \quad (8)$$

### 4.2. Algorithm 2: Detection of the camera obstruction and motion

This algorithm proposed by Deng-Yuan Huang et all in [6] combines three features to detect both camera motion and occlusion.

#### 4.2.1. Background subtraction

It starts by generating the background which will be subtracted from each frame. After that, it calculates the difference between the current image and the background one. If the difference between the current frame and the delayed one is large, it implies the possibility of motion or occlusion of the camera. The difference is designated high if it exceeds a certain threshold τ in case of camera motion when the whole scene appears in binary image. The difference between the current image and the delayed frame is calculated through the following equation:

$$B_D(x,y) = |T_{current}(x,y) - T_{delayed}(x,y)| > \tau \quad (9)$$

$B_D(x,y)$ : Binary image which is the difference between current frame and delayed frame pixel (x, y).
$T_{current}(x,y)$: Intensity of the pixel (x,y) of current frame in grayscale
$T_{delayed}(x,y)$: Intensity of the pixel (x,y) of delayed frame in grayscale.

#### 4.2.2. Second phase: Histogram Analysis

This phase aims to differentiate between normal operation and obstructing the camera. The system compares the number of pixels around the maximum value $f_0$ of the histogram between the current frame and the delayed one, using the following equation:

$$(\sum_{k=-n}^{n} H_{f0+k})_{Frame(n)} \geq (\sum_{k=-n}^{n} H_{f0+k})_{Frame(n-1)} * \theta_{obstruction} \quad (10)$$

$H_{f0+k}$ : Number of intensity level $f_{0+k}$ pixels
$\theta_{obstruction}$ : Detection sensitivity
$[-n, n]$ : Range of intensity levels centred in $f_0$

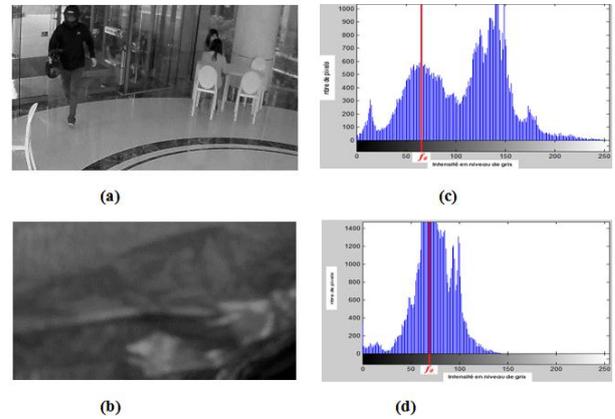

Figure 11 Comparison between the histogram distributions in case of normal operation and in the obstruction of the camera

This phase aims to accurately ensure the detection of the event and reduce the possibility of false alarms. As the occlusion of the camera causes the full or partial disappearance of the captured scene, specifically the contours and edges that represent these objects, the system uses the Sobel filter to detect edges in each image, then it compares the number of pixels belonging to the contours in the current frame with those of the delayed one according to the following inequality:

$$S_{CTn} \leq S_{CTn-1} \times \theta_{contour} \quad (11)$$

$S_{CTn}$: Number of pixels belonging to current frame contour

$S_{CTn-1}$: Number of pixels belonging to delayed frame contour

$\theta_{contour}$ : Detection sensitivity

### 4.3. Algorithm 3: Camera defocus detection

This algorithm is based on the DCT technique. The loss of sharpness caused by malicious events results from the loss of its details, contours and edges. DCT is adapted to this kind of problem because the change in the degree of visibility is reflected by the distribution of the cosine spectrum. In the figure below, the cosine spectrum is clearly different in case of a clear image and a blurred image, specifically in the area of high frequencies. The number of zero coefficients in the high frequency cosine spectrum increases with the increase of the image loss.

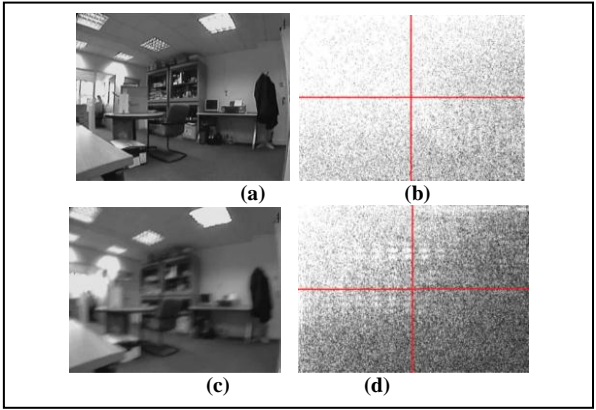

(a)  (b)
(c)  (d)

Figure 12 Comparison between cosine spectra after application

Therefore the number of non-zero coefficients in the high frequency part contained in the cosine spectrum is used in the evaluation of the frame sharpness. After applying DCT on the sequences of images caught by the camera, this algorithm compares the number of non-zero coefficients in the high frequency area which is down on the right.

The proposed system considers that the first initial frame is the back of the scene where there is no motion or change of the original scene and the camera view is clear. Then it calculates the value of $Q_{Base}^{HF}$ that represents the number of coefficients of high frequencies in the background.

$Q_t^{HF}$ is the number of non-zero coefficients in the cosine spectrum precisely in the box which is down on the right. $Q_t^{HF}$ is calculated for each frame at time t using the following equation:

$$Q_t^{HF} = \sum_{y=M/2}^{M-1} \sum_{x=N/2}^{N-1} s(x,y) \qquad (12)$$

$$- s(x,y) = \begin{cases} 1, & C(x,y) > 0 \\ 0, & otherwise \end{cases} \qquad (13)$$

- $C(x,y)$ : cosine spectrum coefficient of pixel (x, y)
- **M and N :** Respectively the height and width of the image of the cosine spectrum.

NB: $Q_t^{HF}$ is calculated only in the high frequency region

The next step is to compare $Q_t^{HF}$ of the current frame with the one of initial frame: $Q_{Base}^{HF}$. If $Q_t^{HF}$ is less than **70%** of $Q_{Base}^{HF}$ then we deduce that the camera is out of focus.

The comparison leads to the following inequality:

$$Q_t^{HF} < \alpha_L \quad \text{and} \quad \alpha_L = Q_{base}^{HF} x \beta_L$$

$\beta_L$ : Detection sensitivity

### 4.4. Algorithm 4: Another camera defocus detection

We implemented a 4[th] algorithm using the FFT *with the same concept and approach as the 3[rd] algorithm* using DCT, in order to estimate the event of camera defocus. But the difference between them is that the last algorithm requires browsing the whole image of the spectrum to extract the number of high frequency coefficients, while via the method based on the DCT we just browse the $\frac{1}{4}$ of the spectrum image where it was locating the high frequencies area.

### 4.5. Algorithm 5: camera Tampering detection

This method described in [11] is adopted for all attempts of surveillance camera tampering. It concentrates processing on the contours and edges of the image because they are the most sensitive to all cases of malfunction of the camera.

The influences on the pixels of the contours are as follows:

- In case of obstruction: total or partial disappearance of the pixels belonging to the contour.

- In case of defocus: degradation of the contour.

- In case of motion: the pixels belonging to the contour change positions precisely those which belong to the background of the scene.

To develop this method, the first step consists in detecting the contour of consecutive frames by applying an optimal thresholding in order to eliminate noises caused by the lighting conditions and to keep the most stable pixels in terms of level of illumination intensity.

The second step counts the number of pixels belonging to the detected contour of the current frame $T_n$:

$$N = \sum_{i=1}^{N} \sum_{j=1}^{M} p(x,y) \quad (14)$$

$$p(x,y) = \begin{cases} 1, & if\ Tn(i,j) = 1 \\ 0, & otherwise. \end{cases} \quad (15)$$

**M and N :** Respectively the height and width of the image

The final step consists of the comparison of the number of edge pixels of the current frame $T_n$ and the delayed frame $T_{n-1}$. If $N_d$ is 130% greater than **N**, then it means that most of the pixels belonging to the contour of the current frame are changing in terms of intensity in the case of loss, reduced visibility or camera motion.

$$N_d = \sum_{i=1}^{N} \sum_{j=1}^{M} d \quad (16)$$

$$d = \begin{cases} 1, & if\ T_n(x,y) \neq T_{n-1}(x,y) \\ 0, & if\ T_n(x,y) = T_{n-1}(x,y) \end{cases} \quad (17)$$

The system compares $N_d$ and N according to the following inequality:
$N_d > \alpha\ x\ N$ , α : Detection sensitivity
In next section, we will expose a new method for the camera tempering detection.

## 1. Proposed Method

Our proposed method for the camera tampering detection proposes a combination of the algorithm 5, algorithm 2 and algorithm3 to detect different type of camera tampering.
The figure 14 shows the flowchart of our proposed method.
The proposed algorithm is tested on several videos recorded in different locations outside and inside. Evaluation criteria are in terms of accuracy, error, and detection sensitivity with variant conditions in the time such as the motion of objects in the scene, and slow or sudden lighting degradation.
Finally, we show the performance comparison shows the superiority of our proposed method.

## 2. Experimental results and discussion

To evaluate Camera Tampering detection algorithms,. We have used 41 video clips. We recorded many video clips with durations between 2.5 minutes and 1 hour using a static camera of 640x480 resolutions. These videos are saved inside enclosed areas (bank, company etc ...) and outside (parking, garden, highway....etc). The scenes contain people and moving objects. And we abruptly changed the brightness and slowly to see the behavior of the different systems. We also have downloaded some videos from youtube containing various attempts of tampering of the surveillance camera. We have implemented the different algorithms using Matlab.
The table1 present the result of the comparison of some used methods for the camera tampering detection.

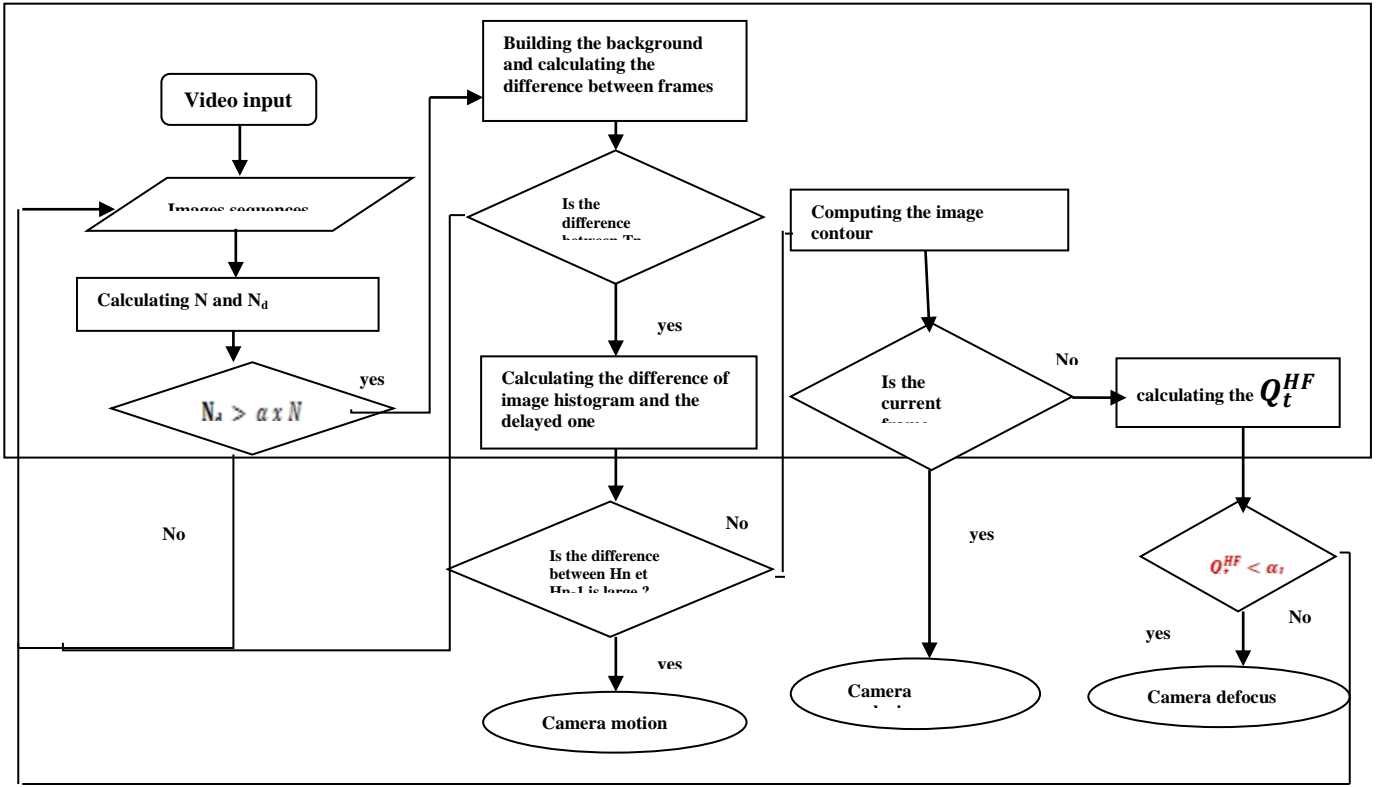

Figure 13 Flowchart of the proposed method

| Algorithm | false detection rate | True detection rate |
|---|---|---|
| Algorithm 1 | 28.5% | 71.5% |
| Algorithm 2 | 16.6% | 84.6% |
| Algorithm 3 | 12.5% | 87.5% |
| Algorithm 4 | 12.5% | 7 (87.5%) |
| Algorithm 5 | 4.8% | 95.,2% |
| Our method | 2.8% | 97.2% |

Table1. Comparison of algorithms performances

The indicated algorithms bellow present several anomalies such as the algorithm 1 detect the motion of big object which covers the camera lens as camera occlusion. Also the simultaneous moving of different objects can disturb the algorithm2 working. In fact, many pixels will be non-zero in the subtraction of the frames and their number will reach the approximate threshold $\theta_B * N_T$ and the system will produce false camera motion detection. Our experiments show the superiority of our method comparing to others.

## 3. Conclusion

Recently the use of video surveillance systems is widely increasing. Different places are equipped by camera surveillances such as hospitals, schools, airports, museums and military places in order to ensure the safety and security of the persons and their property. It becomes significant to guarantee the proper working of these systems. The camera Tampering Detection uses several techniques based on image processing and computer vision. This paper conducts a comparative study of mainly used camera tampering detection algorithms. We have also proposed a new method to detect different type of camera sabotage.